%% file: nips_2018.tex
\documentclass{article}

\usepackage{cite}
\usepackage{amsmath,amssymb,amsfonts}
\usepackage{algorithmic}
\usepackage{graphicx}
\usepackage{textcomp}
\usepackage{xcolor}
\usepackage{multirow}
\usepackage{color}
\usepackage{soul}
\usepackage[normalem]{ulem}
\usepackage{xspace}
\usepackage{tabularx}
\usepackage{array}
\usepackage{booktabs}
\usepackage[preprint,nonatbib]{nips_2018}

\usepackage[utf8]{inputenc} % allow utf-8 input
\usepackage[T1]{fontenc}    % use 8-bit T1 fonts
\usepackage{hyperref}       % hyperlinks
\usepackage{url}            % simple URL typesetting
\usepackage{booktabs}       % professional-quality tables
\usepackage{amsfonts}       % blackboard math symbols
\usepackage{nicefrac}       % compact symbols for 1/2, etc.
\usepackage{microtype}      % microtypography

\title{Can WiFi Estimate Person Pose?}

\author{Fei Wang\textsuperscript{1,2}\thanks{Work done when at CMU.}~~~~Stanislav Panev\textsuperscript{2}~~~~Ziyi Dai\textsuperscript{1}~~~~Jinsong Han\textsuperscript{3} ~~~~Dong Huang\textsuperscript{2}
\\
\small{\textsuperscript{1}Xi'an Jiaotong University~~~~\textsuperscript{2}Carneige Mellon University~~~~\textsuperscript{3}Zhejiang University}\\
\small\texttt{feiwang@cmu.edu, spanev@cmu.edu, dzy219@gmail.com}\\ \small\texttt{hanjinsong@zju.edu.cn, donghuang@cmu.edu}
}
\begin{document}

\maketitle

\begin{abstract}
% Sensing human with WiFi attracts blooming attention because of its distinct properties, such as prevalence, through-wall and resilience to lighting condition. To date, 
WiFi human sensing has achieved great progress in indoor localization, activity classification, etc.
Retracing the development of these work, we have a natural question: can WiFi devices work like cameras for vision applications? In this paper
We try to answer this question by exploring the ability of WiFi on estimating single person pose.
 We use a 3-antenna WiFi sender and a 3-antenna receiver to generate WiFi data. Meanwhile, we use a synchronized camera to capture person videos for corresponding keypoint annotations. 
We further propose a fully convolutional network~(FCN), termed WiSPPN, to estimate single person pose from the collected data and annotations.
Evaluation on over 80k images~(16 sites and 8 persons) replies aforesaid question with a positive answer.~\textit{Codes have been made publicly available at~\url{https://github.com/geekfeiw/WiSPPN}}.

\end{abstract}

\input{tex/introduction}

\input{tex/related.tex}

\input{tex/background.tex}

\input{tex/method.tex}

\input{tex/evaluation}

\section{Conclusion}
In this paper, we build a system and propose a novel network termed WiSPPN for a fine-grained WiFi sensing, i.e., single person pose estimation. The experimental results show that WiFi sensors can achieve a comparable performance in fine-grained human sensing to cameras.

\section*{Acknowledgments}
We thank Jianwei Feng, Zeyi Huang and Sanping Zhou for valuable discussions. Fei Wang is supported by China Scholarship Council.

{\small
\bibliographystyle{IEEEtran}
\bibliography{bibo}
}

\end{document}

%% file: tex/introduction.tex
\section{Introduction}\label{sec:intro}

% Human sensing technology plays a critical role in many applications such as human-computer interaction~\cite{}, health-care system~\cite{} and surveillance control~\cite{}. To meet specific requirements, sensors for human sensing vary a lot like cameras, speakers, radars and IMUs. 
% Recent years, we witness the fast development and ubiquitous deployment of WiFi networks. 
The key components of ubiquitous WiFi networks, WiFi devices, have been widely explored in many human sensing work such as indoor localization~\cite{vasisht2016decimeter,kotaru2015spotfi,qian2018widar2, li2016dynamic} and activity classification~\cite{wang2015understanding,virmani2017position,kotaru2017position}. Retracing the development of these work, a natural question arises: whether WiFi devices can work like cameras for fine-grained human sensing task such as the person pose estimation. If the answer is yes, WiFi could be an alternative or supplementary solution for cameras in some situation such as sensing through-wall, under occlusion and in the dark. Besides the advancement in the physical properties comparing to cameras, WiFi devices are prevalent, requiring less cost in deployment, and rise less privacy concerns for the public.

\begin{figure}[t]
    \centering
    \includegraphics[width=0.88\linewidth]{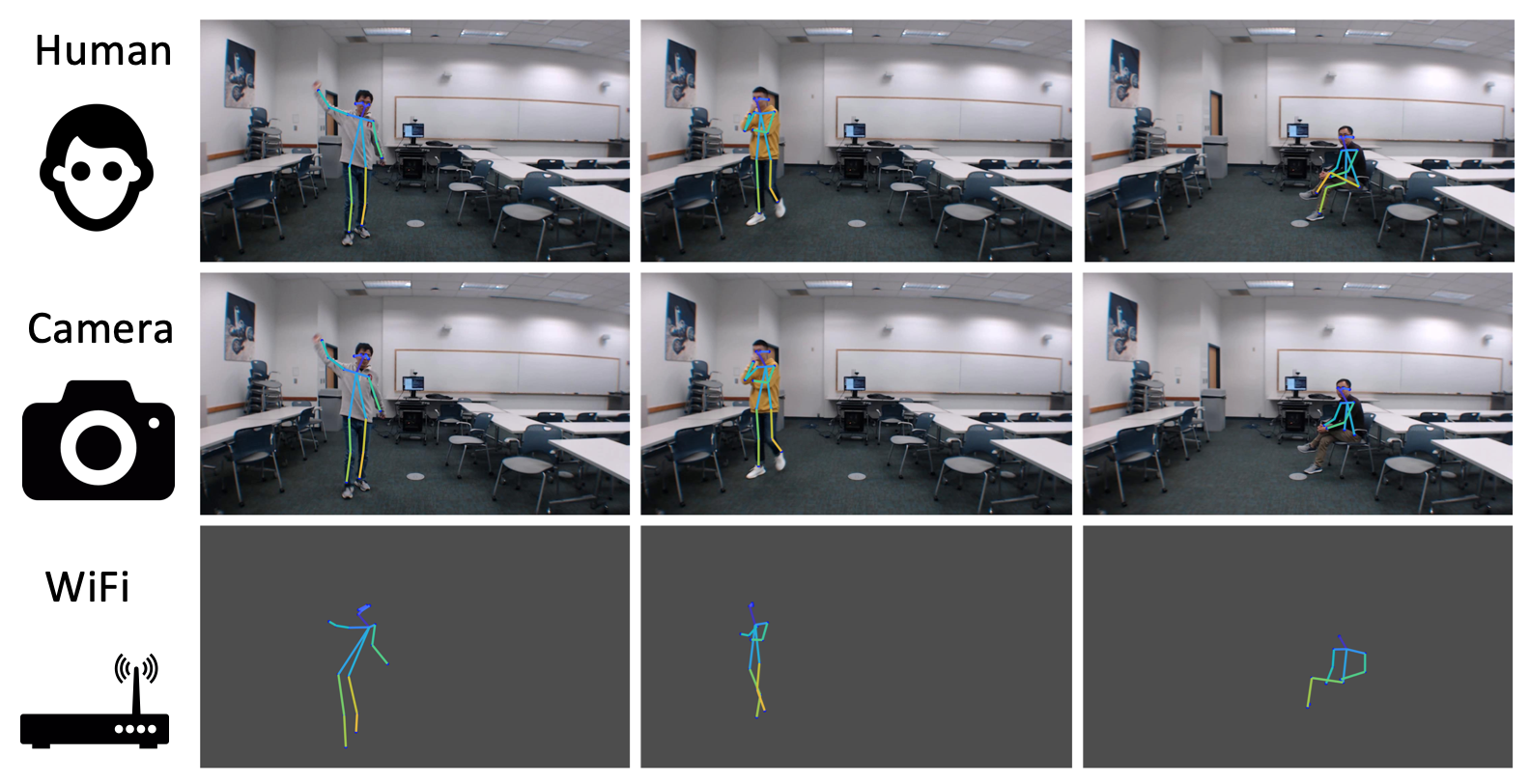}
    \caption{Person pose estimation examples of a camera-based approach~(AlphaPose~\cite{fang2017rmpe}) and the WiFi-based approach~(ours). Rendered images in the~\textit{1st} row are manually marked.}
    \label{fig:first}
\end{figure}

Though estimating person pose estimation with WiFi is with high practical impact for above explanations, it is full of challenges. First, WiFi is designed for wireless communication, which carries no direct information on the person keypoint coordinates. We cannot benefit from the most popular person pose estimation schema in computer vision techniques, inferring the person location from a image then regressing the keypoint heatmaps~\cite{fang2017rmpe,chen2018cascaded}. Thus in order to learn the mapping from WiFi signals to person pose, pose supervision must be prepared and it must be corresponding with WiFi signals. To deal with this problem, we combine a camera with WiFi antennas to capture person videos. The camera and WiFi are synchronized with Unix time to guarantee the correspondence. The pose supervision is derived from videos through the AlphaPose~\cite{fang2017rmpe}, an accurate yet fast open-source person pose estimation repository. 

Second, it is very pioneering that estimating person pose from WiFi signals, thus we have little work to refer. Even in computer vision community, it takes decades to reach an acceptable performance for image or video inputs. 
Generally, deep networks that estimates with WiFi should be  completely different.
After abundant survey, we propose WiSPPN~(abbreviation of WiFi single person pose networks), which is a selective combination of CSI-Net~\cite{wang2018csi}, ResNet~\cite{he2016deep} and FCN~\cite{long2015fully}. To be exact, we utilize the up-sampling stage of CSI-Net to encoding WiFi signals. Then we use ResNet to extracting feature. Moreover, we propose an innovative pose embedding approach which is inspired by the adjacency matrix in the graph theory. This approach would take the length constraint of pose coordinates and make pose estimation can be done with FCN.

When solve these two challenges, we achieve single person pose estimation with WiFi. Evaluation over 80k images shows that our approach achieve single person pose well.

The contribution of this paper can be summarized as follows.

1. We put forwards a question that whether WiFi can be used like cameras for vision problem. We positively answered this question by demonstrating that WiFi singles can be used for single person pose estimation.

2. To answer this question, we built a multi-modality system, collected a dataset and propose a novel deep networks to learn the mapping from WiFi signals to person keypoint coordinates. 

% 3. We experimentally evaluated our approach over 80k images.
% We tested its generalization ability in time, person ID and test...

% 3. We broadened the WiFi sensing range with a fine-grained and challenging task, pose estimation. Our approach may inspire and encourage researchers to do fine-grained human sensing with other ubiquitous sensors.

% The idea of FCN inspires us that 
% Still now, pose estimation is still an open problem and attracts lots of attention because it is core competition in COCO and Mapillary Recognition Challenge, one of the most significant competition in computer vision community~\footnote{http://cocodataset.org/\#keypoints-2018}.
% a future technology    
% We witness the development of human

%% file: tex/related.tex
\section{Related Work}\label{sec:related}

\textbf{Camera.}~Estimating multi-person pose from RGB images is a widely-studied problem~\cite{wei2016cpm,cao2017realtime,newell2016stacked}. The leading solutions of COCO Challenge, such as AlphaPose~\cite{fang2017rmpe} and CPN~\cite{chen2018cascaded}, are prone to apply a person detector to crop every person from images, then to do single person pose estimation from cropped feature maps, regressing the heatmap of each body keypoints. Coordinates with highest confidence are the estimation of single person pose. 
% Despite RGB images, images captured by depth camera~\cite{}, infrared camera~\cite{} are also be studied for estimating person pose. 

\textbf{Other sensors.}~Due to the potential usages of pose estimation, researchers have applied many other sensors to estimate person body pose or sketch.  Wall++~\cite{zhang2018wall++} enables a common wall large electrodes with water-based nickel~\cite{water_based} painted. Then the Wall++ can sense airborne electromagnetic variance caused by human body and estimate person pose. LiSense~\cite{li2015human} and StarLight~\cite{li2016practical} use ceiling LED and photo-resistor blanket to capture human body shadow on the blanket, then reconstruct body sketch/pose. With promoted technology, person's hand can also be reconstructed by similar systems~\cite{li2017reconstructing}. RF-Capture~\cite{adib2015capturing} and RF-Pose~\cite{zhao2018through} implements radars with frequency modulated continuous wave~(FMCW) equipment to estimate person body sketch/pose. Even, single-photon sensors can be used to reconstruct person body~\cite{shin2016photon,o2017reconstructing}. Comparing to these sensors, devices with WiFi chips may be the most pervasive, such as routers, cell phones and blooming Internet-of-Things.

%% file: tex/background.tex
\section{Background}~\label{sec:background}

\begin{figure}[t]
    \centering
    \includegraphics[scale=0.6]{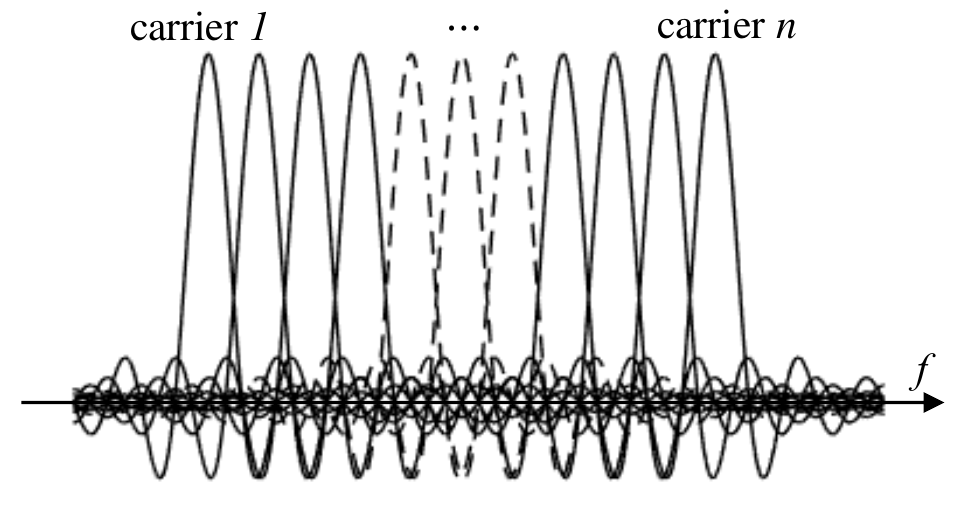}
      \includegraphics[scale=0.6]{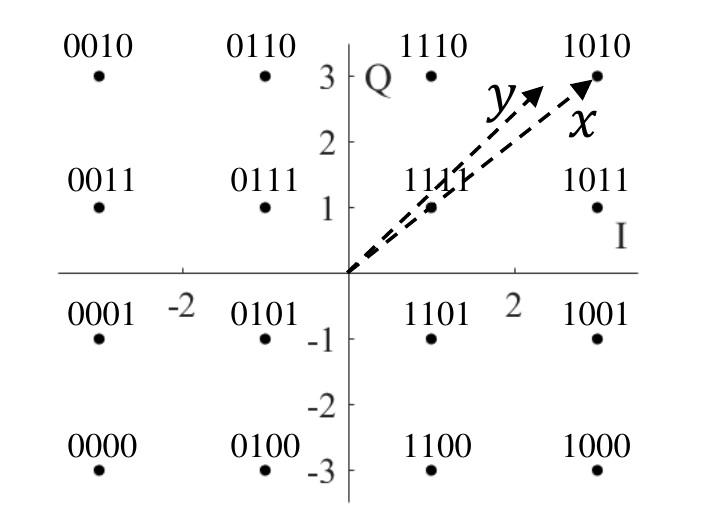}
    % \vspace{-10pt}
    \caption{Left: OFDM. information is carried by $n$ carriers; Right: 16-QAM. for each carrier, one modulated signal carries 4 bits data. If carrying `1010', it would be modulated to $x=3+3i$ and broadcasts. The received signal is $y$, and the variation during propagation is $h=y/x$, which is used for human sensing.}
    \label{fig:wifi}
\end{figure}

\subsection{WiFi Signals and Channel State Information}\label{sec:wifi}
Under IEEE 802.11 n/g/ac protocols, WiFi works around 2.4/5GHz~(central frequency) with multiple channels. In each channel, the bandwidth is 20/40/80/160MHz. Within the band, carriers with different frequencies are modulated to carry information for wireless communication in parallel, which is called orthogonal frequency division multiplexing~(OFDM) and illustrated in the left of Fig.~\ref{fig:wifi}. During propagation, WiFi carriers decay in power and shift in phase. Moreover, their frequencies may also change when encountering a moving object due to the Doppler Effect. Channel State Information~(CSI), a physical layer indicator, can be used to represent these variation of carriers. 

Take the modulation method of 16-quadrature amplitude modulation~(16-QAM) for example~\footnote{More modulation method can be found at, http://mcsindex.com/}, as shown in the right of Fig.~\ref{fig:wifi}, one modulated carrier contains 4bits information one time. When the sender sends a `1111' to the receiver, the carrier would be modulated to $x=1+1i$. If the receiver receives a $y=0.8+0.9i$. Thus the variation happening during propagation is $h=y/x=0.2+3.4i$, which is called CSI of this carrier. For the human sensing application, human body as an object, is able to make carrier change. In this paper, we aims to learn the mapping rule from the change to single person pose coordinates. We set WiFi working within a 20MHz band, the CSI of 30 carriers can be obtained through a open-source tool~\cite{halperin2011tool}. In the remaining content of this paper, WiFi signals and CSI indicate the same thing if not stated specially.

\subsection{AlphaPose}~\label{sec:alphapose}

AlphaPose is an open-source multi-person pose estimation repository~\footnote{https://github.com/MVIG-SJTU/AlphaPose/tree/pytorch}, which is also applicable for single person pose estimation. AlphaPose is a two-step framework, which first detects person bounding boxes by a person detector~(YOLOv3~\cite{redmon2018yolov3}) then estimates pose for each detected box by the pose regressor. With the innovative regional multi-person pose estimation framework~(RMPE)~\cite{fang2017rmpe}, AlphaPose gains  estimation resilience to the inaccurate person detection, which largely facilitates the pose estimation performance.  Please refer~\cite{fang2017rmpe} for more details on AlphaPose and RMPE. 

When applied to single person pose estimation, AlphaPose generates $n$ three-element predictions in the format of $(x_i,y_i;c_i)$, where $n$ is the number of keypoints to be estimated, $x_i$ and $y_i$ are the coordinates of the~\textit{i-th}~keypoint, and $c_i$ is the confidence of the above coordinates. In this paper, we use the COCO person keypoint setting and $n$ is 18. Four estimation examples of AlphaPose are shown in the top of Figure.~\ref{fig:first}.

% \begin{table*}[t]
% \centering
% \begin{tabular}{|l|l|l|l|l|l|l|l|l|}
% \hline
% 1: Nose        & 2: Neck         & 3: R. Shoulder & 4: R. Elbow & 5: R. Wrist & 6: L. Shoulder & 7: L. Elbow & 8: L. Wrist & 9: R. Hip \\ \hline
% 10: R. Knee & 11: R. Ankle & 12: L. Hip      & 13: L. Knee  & 14: L. Ankle & 15: R. Eye    & 16: L. Ear  & 17: R. Ear & 18: L. Ear \\ \hline
% \end{tabular}
% \caption{Body keypoint ordering. R. and L. stand for Right and Left, respectively.}
% \end{table*}

% ~\footnote{https://github.com/MVIG-SJTU/AlphaPose/blob/pytorch/doc/output.md\#keypoint-ordering}

%% file: tex/method.tex
\section{Methodology}~\label{sec:methodology}

\subsection{System Build}\label{sec:system}
\begin{figure}[t]
    \centering
    \includegraphics[width=0.8\linewidth]{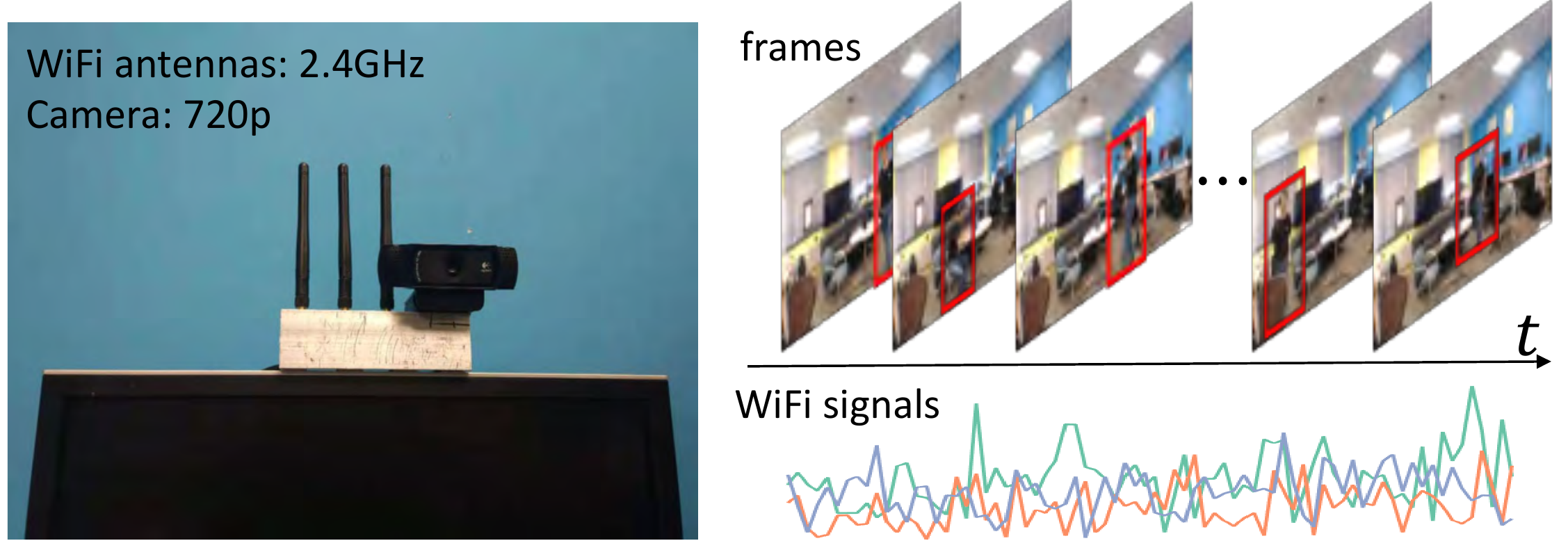}
    \caption{The receiver. Left: A camera is aligned with 3 antennas and records video frames along with WiFi. Right: Examples of frames and WiFi signals of the 10th,20th and 30th carriers.}
    \label{fig:system}
\end{figure}

To do pose estimation from WiFi by learning, we must have pose annotations. However we cannot mark person pose coordinates in the WiFi signals, thus we use a camera aligned with WiFi antennas to capture person videos. Then the video is processed by AlphaPose~\cite{fang2017rmpe} for pose annotations in coordinates and confidences. Besides, the camera and WiFi antennas are synchronized by the their recorded time-stamps. The WiFi CSI recording system is comprised with 2 ends, one 3-antenna sender and one 3-antenna receiver. The sender broadcasts WiFi signals, meanwhile the receiver parses CSI through~\cite{halperin2011tool} when receiving the broadcasting WiFi. In our setting, the parsed CSI is a tensor with the size of $n\times 30 \times 3\times 3$, where the $n$ is for the number of received WiFi packages; 30 is for the subcarrier number; the last two 3s represent the antenna numbers of sender and receiver, respectively. The WiFi pose system is shown Fig.~\ref{fig:system}. In our data acquisition, we set the sampling rate of WiFi devices and the camera as 100Hz and 20Hz, respectively. Thus we have a paired dataset in which every 5 CSI samples and one image frame are synchronized by their time-stamps.

\begin{figure}[t]
    \centering
    \includegraphics[width=0.6\linewidth]{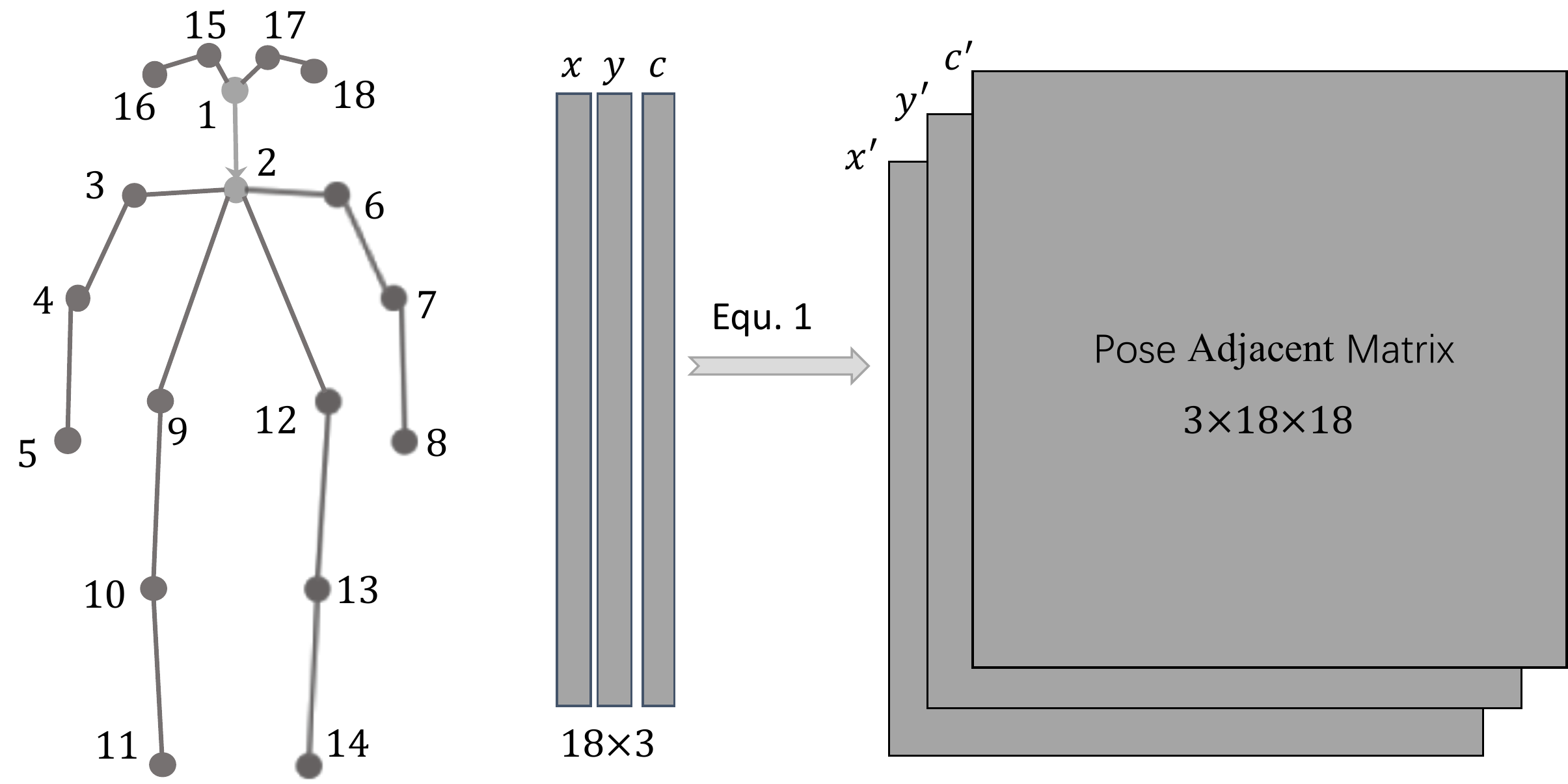}
    \caption{CMU keypoint ordering~\cite{cao2017realtime} and pose adjacent matrix.}
    \label{fig:pam}
\end{figure}

\begin{figure*}[t]
    \centering
    \includegraphics[width=1.0\linewidth]{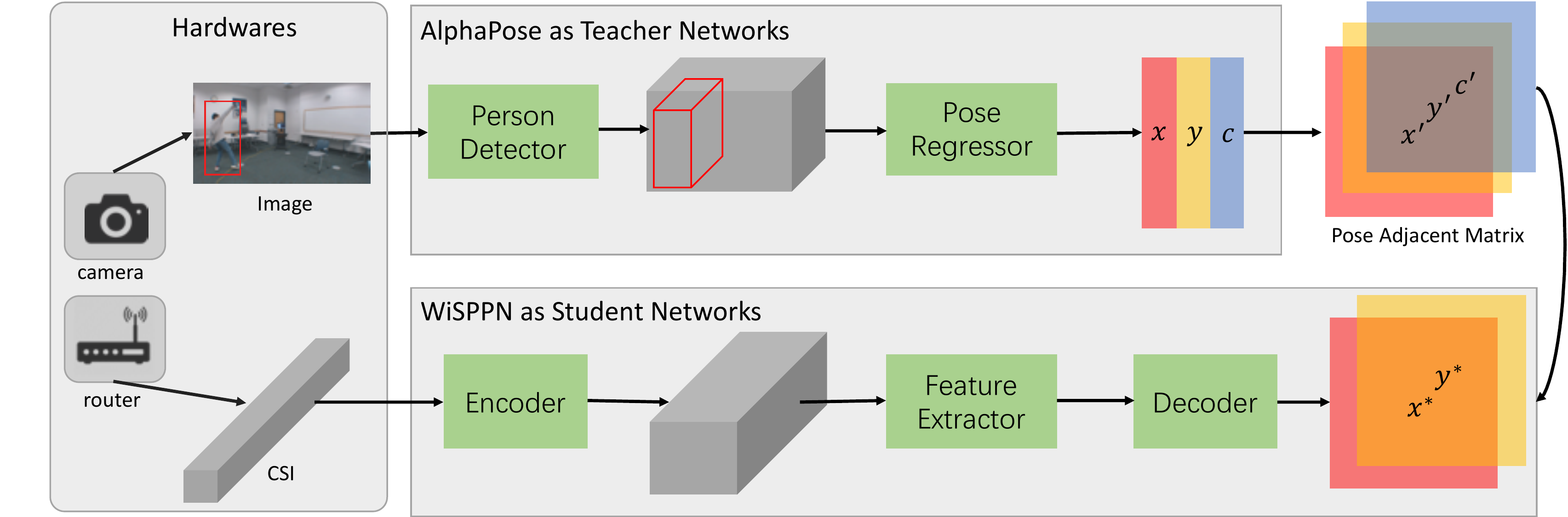}
    \caption{WiSPPN System Framework}
    \label{fig:framework}
\end{figure*}

\subsection{Pose Adjacent Matrix}\label{sec:pam}
As Section.~\ref{sec:alphapose} said, we have 18 person keypoint coordinates, $(x,y;c)$, for each video frame from those with single person. Note that AlphaPose may predict multiple persons for single-person frame~(false-positive), in this situation, we only keep the one with highest confidence. Many work have demonstrated that regressing keypoint coordinates harms the generalization ability in person pose estimation~\cite{}.
Thus in this paper, we learn to regress pose adjacent matrix~(PAM), instead of directly regressing person keypoint coordinates. The PAM is a $3\times 18 \times 18$ matrix, also annotating the pose coordinates and confidences of 18 keypoints. The PAM is comprised of 3 submatrixes, $x'$, $y'$ and $c'$. The $x'$ and $c'$ are generated by Equ.~(1) from the 18 three-element entries: $(x_i,y_i;c_i),i \in [1,2,...,18]$. The $y'$ is generated similar to generating  $x'$. 

\begin{equation}
x_{i,j}' = \left\{\begin{matrix}
x_i - x_j, & i\neq j; \\
x_i, &i=j.
\end{matrix}\right.
~~~~
c_{i,j}' = \left\{\begin{matrix}
c_i \times c_j, & i\neq j; \\
c_i, &i=j.
\end{matrix}\right.
\label{equ:pam}
\end{equation}

To be specific in the Graph Theory view, we take person skeleton as a directed complete graph~(DCG)~\cite{west1996introduction}, each keypoint as a node of the graph. For the $x'$ and $y'$ of PAM, the diagonal items are the coordinate values in $x$ and $y$ axes of these 18 nodes, respectively. Meanwhile, the elements in other indexes are the displacement of two adjacent nodes--hence the name of pose adjacent matrix. For the $c'$ of PAM, the diagonal items are the confidence values of corresponding nodes. While we think the displacement between two nodes happens independently, thus we computer $c_{i,j}'=c_i \times c_j$ for other indexes. Finally, we innovatively embed person keypoint coordinates as well as the displacements between keypoints into the PAM. 

 The main advancement of PAM is that it provides additional constraint of human skeleton shape for person pose estimation. Take the displacement in $y$ axis from the~\textit{nose} to the~\textit{neck} for example, the displacement is a negative value in majority of situation because the neck is below the nose in human skeleton for a standing person. The negativity take the direction of nose to neck as constraint. Besides, the absolute value of the displacement take the length of nose to neck as constraint. When we regress PAM, the additional regression on its displacements work as the regularization item, taking person skeleton shape into consideration and highly increasing the approach generalization ability comparing to regressing keypoint coordinates directly. 
%  Excluding above qualitative analysis, we will experimentally study this comparison with an ablation evaluation in Section.~\ref{sec:eval}. 

\subsection{Network Framework}
We donate the training dataset as $\mathbf{D}=\{( \mathrm{I}_t, \mathrm{C}_t), t \in [1, n] \}$, where $\mathrm{I}_t$ and $\mathrm{C}_t$ are a pair of synchronized image frame and CSI series, respectively; $t$ means the sampling moment; and $n$ is the dataset size. We propose a novel deep network to train $\mathbf{D}$ for the purpose of learning to a mapping rule from CSI series to person body keypoints. The network framework is comprised of  AlphaPose~\cite{fang2017rmpe} as a teacher network and  WiSPPN as the student network, shown in Fig.~\ref{fig:framework}. 
The teacher and student network are termed as $\mathbf{T}(\cdot)$ and $\mathbf{S}( \cdot)$, respectively. For each $( \mathrm{I}_t, \mathrm{C}_t)$ pair,  $\mathbf{T}(\cdot)$ takes $\mathrm{I}_t$ as input, and outputs the corresponding body keypoint coordinates and confidence, $(x_t, y_t; c_t)$, with the person detector and pose regressor. We then convert the outputs to a body pose adjacent matrix, $\mathrm{PAM}_t$, with aforesaid Euqation.~\ref{equ:pam}. We formulize the operation of the teacher network as $\mathbf{T}(\mathrm{I}_t) \rightarrow  \mathrm{PAM}_t$, where $\mathrm{PAM}_t$ is the cross-modality supervision to teach $\mathbf{S}(\cdot)$. 

We go into details on $\mathbf{S}(\cdot)$, i.e., WiSPPN. In the training stage, $\mathbf{S}(\cdot)$ takes  $\mathrm{C}_t$ as input, and outputs a corresponding prediction of pose adjacent matrix. Then $\mathbf{S}(\cdot)$ is optimized by the supervision of $\mathrm{PAM}_t$. As shown in Fig.~\ref{fig:framework}, WiSPPN consists of three key modules, i.e., the encoder, feature extractor and the decoder. A $\mathrm{C}_t$ is converted to the $\mathrm{PAM}_t$ prediction undergoing these three modules successively. Next we explain our designing intentions and parameter details on these three modules.

\begin{figure*}[t]
    \centering
    \includegraphics[width=1\textwidth]{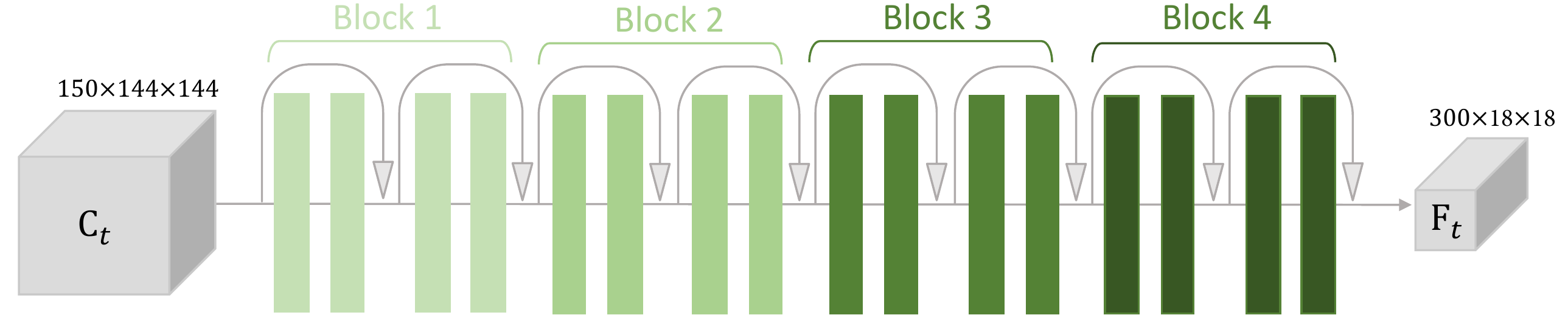}
    \caption{Four stacked residual blocks~(16 convolutional layers) work as the feature extractor, and convert $\mathrm{C}_t$ to $\mathrm{F}_t$.  }
    \label{fig:residual}
\end{figure*}

1. Encoder. The encoder is designed to upsample $\mathrm{C}_t$ to a proper width and height which are suitable for the mainstream convolutional backbone networks such as VGG~\cite{simonyan2014very} and ResNet~\cite{he2016deep}. Recall that our WiFi system is comprised of a sender and a receiver both with 3 antennas, which outputs CSI samples with size of $30\times 3\times 3$ through a open-source tool~\cite{halperin2011tool}, where the 30 is the number of OFDM carriers described in Section.~\ref{sec:wifi}. 
As said in Section.~\ref{sec:system}, one image matches with 5 continuous CSI samples due to the sampling rate inconformity, leading to $\mathrm{C}_t \in \mathbf{R}^{ 5\times30\times 3\times 3}$, and we reshape it to be $150\times3\times3$ along the time axis, which makes  $\mathrm{C}_t \in \mathbf{R}^{ 150\times 3\times 3}$.
However, a general RGB image is with size like $3 \times 224\times 224$, where 3 is for the 3 color channels in Red, Green and Blue; and 224s are the height and width of the image. To enlarger the width and height of CSI samples, CSI-Net~\cite{wang2018csi} use 8 stacked transposed convolutional layers to upsample its input from size of $30\times 1\times 1$ to $6\times 224\times 224$ gradually, which is operation-consuming. In WiSPPN, we apply one bilinear interpolation operation to directly convert $\mathrm{C}_t \in \mathbf{R} ^{150 \times 144 \times 144} $ for further feature extraction.

2. Feature extractor. With the upsampled $\mathrm{C}_t$, the feature extractor are used to learn efficient features for the person pose estimation. Because  $\mathrm{C}_t$ lacks spatial information of person body keypoints comparing to images, we need a powerful feature extractor to release its spatial information. Conventionally, a deeper network could have a more powerful feature learning ability. Thus we tend to use a deeper network as the feature extractor of WiSPPN. However, deeper networks are prone to be gradient vanishing or gradient exploding because the chain rule in the backpropagation optimization could result in exponential gradients in the very deep convolutional layers. The ResNets~\cite{he2016deep} are a cluster of the most widely-used backbone networks in deep learning domain, especially in the computer vision research. The ResNets alleviate this problem by the shortcut connection and residual blocks. Considering this advantage, we stack 4 basic blocks of ResNet~\cite{he2016deep}~(16 convolutional layers) as the feature extractor of WiSPPN~(shown in Fig~\ref{fig:residual}), which learn features with a size of $300\times 18\times 18$, termed as $\mathrm{F}_t$. 
The detailed parameters of feature extractor are listed in Table.~\ref{tab:residual}. Note that a batch normalization~\cite{ioffe2015batch} and a rectified linear unit activation~\cite{krizhevsky2012imagenet} follow every convolutional layer, successively.

\begin{table}[t]
\centering
\begin{tabular}{|c|c|c|}
\hline
\textbf{Input.}     &   \multicolumn{2}{c|}{$\mathrm{C}_t \in \mathbf{R}^{150\times 144\times 144}$}                             \\\hline\hline

Block name & Output size & Parameters                       \\\hline

Block 1    & 150x144x144 & $ \begin{bmatrix}
3\times3, c=150, s=1  \\ 
 3 \times 3, c=150,s=1 \\
3\times 3, c=150,s=1\\
3\times 3, c=150,s=1
\end{bmatrix} $ \\\hline
Block 2    & 150x72x72 & $ \begin{bmatrix}
3\times3, c=150, s=2  \\ 
 3 \times 3, c=150,s=1 \\
3\times 3, c=150,s=1\\
3\times 3, c=150,s=1
\end{bmatrix} $ \\\hline
Block 3    & 300x36x36 & $ \begin{bmatrix}
3\times3, c=300, s=2  \\ 
 3 \times 3, c=300,s=1 \\
3\times 3, c=300,s=1\\
3\times 3, c=300,s=1
\end{bmatrix} $ \\\hline
Block 4    & 300x18x18 & $ \begin{bmatrix}
3\times3, c=300, s=2  \\ 
 3 \times 3, c=300,s=1 \\
3\times 3, c=300,s=1\\
3\times 3, c=300,s=1
\end{bmatrix} $ \\ \hline \hline 
\textbf{Output.}     &   \multicolumn{2}{c|}{$\mathrm{F}_t \in \mathbf{R}^{300\times 18\times 18}$}                             \\\hline
\end{tabular}
\caption{Parameters of the feature extractor. The $3\times 3$ means a convolutional layer with $3\times 3$ kernel; $c$ and $s$ stand for the~\textit{out-channel} and~\textit{stride} of a convolutional layer.}
\label{tab:residual}
\end{table}

\begin{figure}
    \centering
    \includegraphics[width=0.7\linewidth]{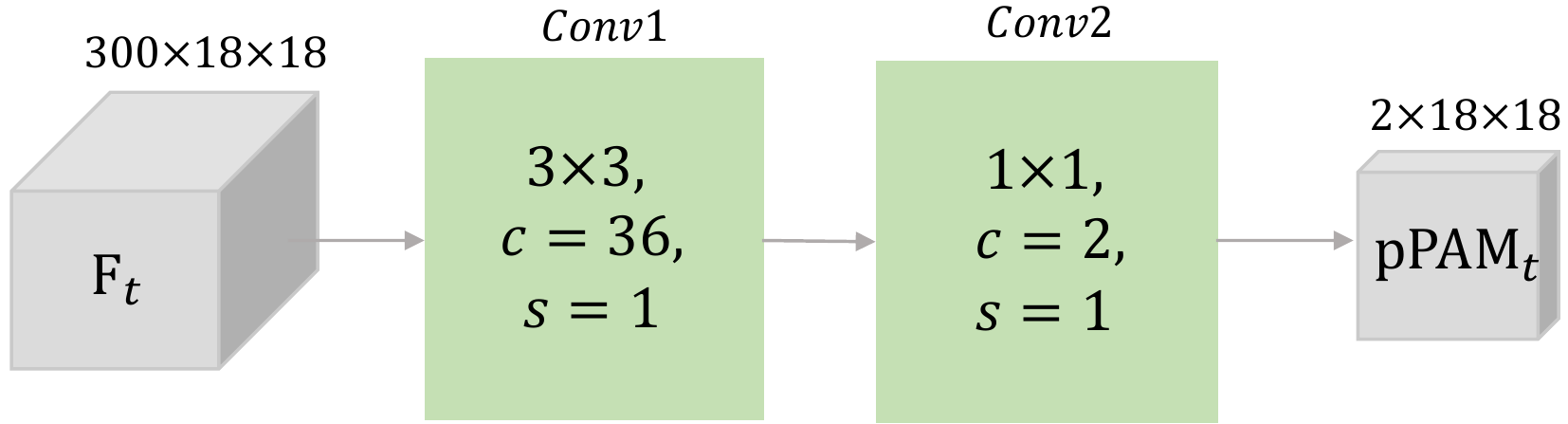}
    \caption{Two convolutional layers work as the decoder to predict pose adjacent matrix. Abbreviations share the same meaning as in Table.~\ref{tab:residual}.}
    \label{fig:decoder}
\end{figure}

3. Decoder. The decoder is designed to do shape adaption between the learned features, $\mathrm{F}_t$, and the supervision outputted by $\mathbf{T}(\cdot)$, $\mathrm{PAM}_t$. As described in Section.~\ref{sec:pam}, the pose adjacent matrix is a novel form for embedding the 18 body keypoint coordinates and the corresponding confidences, and is with the size of $3\times 18\times 18$. In the pose estimation task, a body keypoint can be localized as in two coordinates, i.e., the $x$ axis and the $y$ axis. Thus the decoder is designed to take $\mathrm{F}_t$ as input and predict pose adjacent matrix within $(x,y)$ dimensions, leading to a predicted pose adjacent matrix $\mathrm{pPAM}_t \in \mathbf{R}^{2\times 18\times 18}$. 
To achiever this purpose, we stack two convolutional layers illustrated in Fig.~\ref{fig:decoder}, where $Conv1$ is mainly to release channel-wise information~(from 300 to 36); and $Conv2$ is mainly to further reorganize spatial information of $\mathrm{F}_t$ with the $1\times 1$ convolutional kernels.

Summarily, with the encoder, feature extractor, and decoder, the student network, WiSPPN, predicts a pose adjacent matrix on each CSI input, $\mathrm{C}_t$. We formalize this process as~$\mathbf{S}(\mathrm{C}_t) \rightarrow  \mathrm{pPAM}_t$. During training stage, every predicted  $\mathrm{pPAM}_t$ is supervised by the corresponding result of the teacher network, i.e., $\mathrm{PAM}_t$. 
Once the student network learns well, it gains the ability to do single person pose estimation only with CSI input. Next we describe processes of training stage, including the loss computation and implementation details.

\subsection{Pose Adjacent Matrix Similarity Loss}

As above description, $\mathbf{T}(\cdot)$ outputs $\mathrm{PAM} \in \mathbf{R}^{3\times 18 \times 18}$ as supervisions, and $\mathbf{S}(\cdot)$ outputs $\mathrm{pPAM} \in \mathbf{R}^{2\times 18 \times 18}$ as predictions. With the supervisions and the predictions,  L2 loss is a basic option to be applied to optimize WiSPPN as follows.
\begin{equation}
    \mathcal{L} =  \left \|\mathrm{pPAM}^x-\mathrm{PAM}^x\right \|_2^2 +  \left \|\mathrm{pPAM}^y-\mathrm{PAM}^y \right\|_2 ^2
    \label{equ:l2}
\end{equation}
where $\left \|\cdot \right\|_2^2$ is a operator to compute L2 distance; $\mathrm{pPAM}^x$ and $\mathrm{PAM}^x$ are the prediction and supervision of pose adjacent matrix for body keypoint coordinate in the $x$ axis, respectively; $\mathrm{pPAM}^y$ and $\mathrm{PAM}^y$ are with similar representation while in the $y$ axis. 
% For simplicity, we use $\mathcal{L}^x$ and $\mathcal{L}^y$ to donate $\left \|\mathrm{pPAM}^x-\mathrm{PAM}^x\right \|_2^2$ and $\left \|\mathrm{pPAM}^x-\mathrm{PAM}^x\right \|_2^2$, receptively. Thus Equation.~\ref{equ:l2} becomes  $\mathcal{L} = \mathcal{L}^x + \mathcal{L}^y$.

In this paper, we take the prediction confidence of keypoints in to the loss computing as follows.
\begin{equation}
    \mathcal{L} = \mathrm{PAM}^c*(\left \|\mathrm{pPAM}^x-\mathrm{PAM}^x\right \|_2^2 +  \left \|\mathrm{pPAM}^y-\mathrm{PAM}^y \right\|_2 ^2
    \label{equ:l2})
\end{equation}

% \begin{equation}
%     \mathcal{L} =  \mathrm{PAM}^c*(\left \|\mathrm{pPAM}^x-\mathrm{PAM}^x\right \|_2^2 +  \left \|\mathrm{pPAM}^y-\mathrm{PAM}^y \right\|_2 ^2)
% \end{equation}

% Inspired by a very recent IOU~loss~\cite{yu2016unitbox,rezatofighi2019generalized}

% In the COCO 2018 Keypoint Detection Task, the algorithm is evaluated by~
% % \footnote{\url{http://cocodataset.org/#keypoints-eval}}

% Object keypoint similarity 

% \begin{equation}
%     \mathrm{OKS_i} = \mathrm{exp}({-\frac{(pd_i^x-gt_i^x)^2+(pd_i^y-gt_i^y)^2}{area \times var_i}})
% \end{equation}

% \begin{equation}
%     \mathrm{OKS} =  \frac{ \sum_{i=1}^{K} \mathrm{OKS_i} \times \delta (c_i==1)}{\sum_{i=1}^{K}\delta(c_i==1)}
% \end{equation}

% \begin{equation}
%     area = D(gt_{Rshoulder},gt_{LHip})
% \end{equation}

% For the pose adjacent matrix, we borrow the compuation from OKS and propose pose adjacent matrix similarity~(PAMS).

% \begin{equation}
% \mathrm{PAMS} =  \frac{ \sum_{i,j=1}^{K} \mathrm{PAMS_{(i,j)}} \times c_{(i,j)}'}{\sum_{i,j=1}^{K}c_{(i,j)}'}
% \end{equation}

% \begin{equation}
% \mathrm{PAMS_{(i,j)}} =  \mathrm{exp} ( {-\frac{(pd_{(i,j)}^x-gt_{(i,j)}^x)^2+(pd_{(i,j)}^y-gt_{(i,j)}^y)^2}{area_{(i,j)} \times var_{(i,j)}}})
% \end{equation}

% PAMSLoss = -PAMS

% According 

% top body area to replace 

% \begin{equation}
% area_{(i,j)} = \left\{\begin{matrix}
% D^2(gt_{Rshoulder},gt_{LHip})/2, & i\neq j; \\
% D(gt_{i},gt_{j}), &i=j.
% \end{matrix}\right.
% \end{equation}

% \begin{equation}
% var_{(i,j)} = \left\{\begin{matrix}
% var_i, & i\neq j; \\
% \sqrt{var_i \times var_j}, &i=j.
% \end{matrix}\right.
% \end{equation}

\subsection{Training Details and Pose Association}
We implemented WiSPPN with Pytorch 1.0. The network is trained for 20 epochs with initial learning rate of 0.001, batch size of 32 and Adam optimizer. The learning rate decays by 0.5 at the epoch of 5th, 10th and 15th. 

Once WiSPPN trained, we use it to estimate person pose from testing CSI samples. Taking one sample for example, we can get a predicted PAM~($\mathrm{pPAM} \in \mathcal{R}^{2\times 18\ times 18}$). We take the diagonal elements in $\mathrm{pPAM}$ as the body keypoint prediction by following equations.

For $x$ axis:
\begin{equation}
    x^*_k = \mathrm{pPAM}_{(1,k,k)}, k \in [1,18]
\end{equation}
For $y$ axis:
\begin{equation}
    y^*_k = \mathrm{pPAM}_{(2,k,k)}, k \in [1,18]
\end{equation}.

%% file: tex/evaluation.tex
\section{Evaluation}\label{sec:eval}

\subsection{Data Collection}

% First 80\% frames were as training, the remaining 20\% as test. 
% Leave-one-person out. 

\begin{figure*}[t]
    \centering
    \includegraphics[scale=0.135]{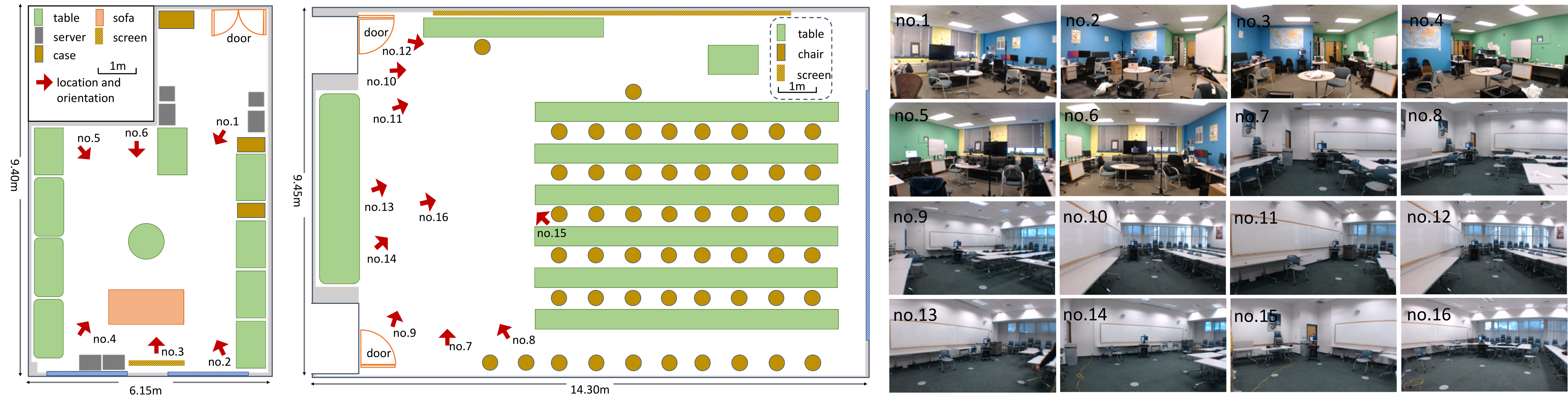}
    \includegraphics[scale=0.135]{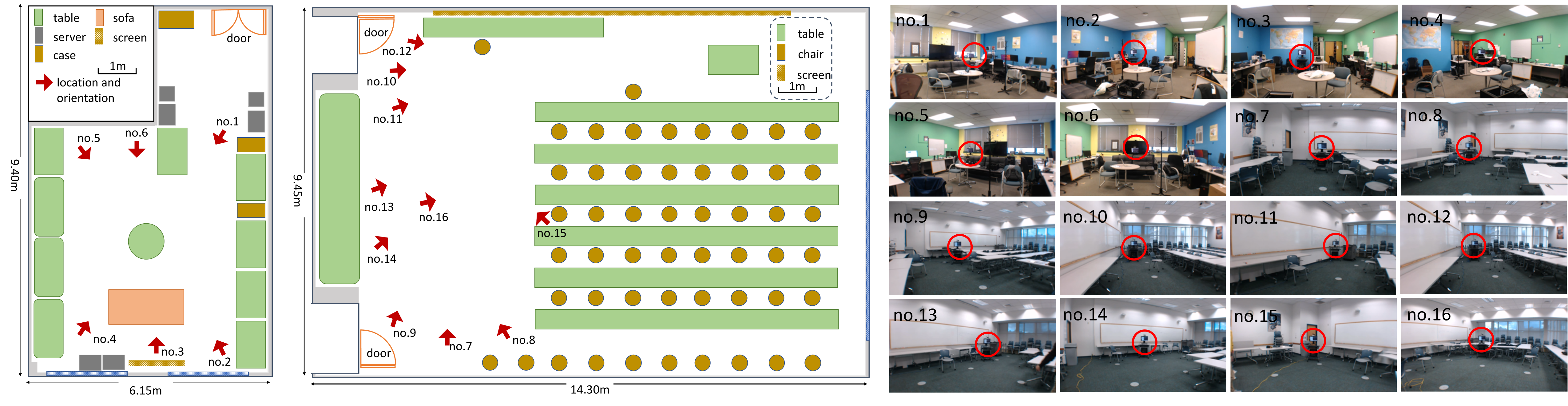}
    \caption{Floor plans and site images of data collection environments. Data were collected at 16 sites from 2 rooms. Arrows mark the location and orientation of WiFi receivers; circles mark the corresponding location of WiFi senders.}
    \label{fig:sites}
\end{figure*}

We collected data under an approval of Carnegie Mellon University IRB~\footnote{No. STUDY2018\_00000352}. We recruited 8 volunteers, and asked them to do casual daily actions in two rooms of the campus, one laboratory room and one class room. Floor plans and data collection positions are illustrated in Fig.~\ref{fig:sites}.  During the actions, we run the system in Fig.~\ref{fig:system} to record CSI samples and videos, simultaneously. For each volunteer, data of his first 80\% recording is used to train the networks, and data of the last 20\% recording is used to test the networks.
The data size of training and testing are 79496 and 19931, respectively.

\subsection{Experimental Results}

Percentage of Correct Keypoint~(PCK) is widely used to evaluate the performance of proposed approach ~\cite{andriluka20142d,yang2013articulated,newell2016stacked}. 
\begin{equation}
    \mathrm{PCK_i@}a =  \frac{1}{N} \sum_{i=1}^{N} \mathrm{I}\left(\frac{\left \|pd_i-gt_i  \right \|^2_{2} }{\sqrt[2]{rh^{2}+lh^{2}}} \leq a \right),
    \label{eq:pck}
\end{equation}
where $\mathrm{I(\cdot)}$ is a binary indicator that outputs 1 while true and 0 while false.
are the same as Equation. $N$ is the number test frames. $i$ denotes the index of body joint and $i \in \left \{1,2,...,18 \right \}$. The $rh$ and $lh$ are for the positions of the right shoulder and  the left hip, respectively. Thus the $\sqrt[2]{rh^{2}+lh^{2}}$ can be regarded as the length of the upper limb, which is used to normalize the prediction error, $\left \|pd_i-gt_i  \right \|^2_{2}$, where $pd$ is prediction coordinates and $gt$ is the ground-truth.

% $\left \|pd_i^p-gt_i^p  \right \|^2_{2}$ is the Euclidean pixel distance between the prediction and ground-truth, which is normalized by the diagonal length of the person bounding box, $\sqrt[2]{w^{p2}+h^{p2}}$. To get person bounding boxes, we aligned body joint coordinates from OpenPose~\cite{cao2017realtime} with the bounding box from Mask R-CNN~\cite{he2017mask} (see Figure~\ref{fig:bbox}). 

\begin{figure}
   \centering
    \includegraphics[width=1.0\linewidth]{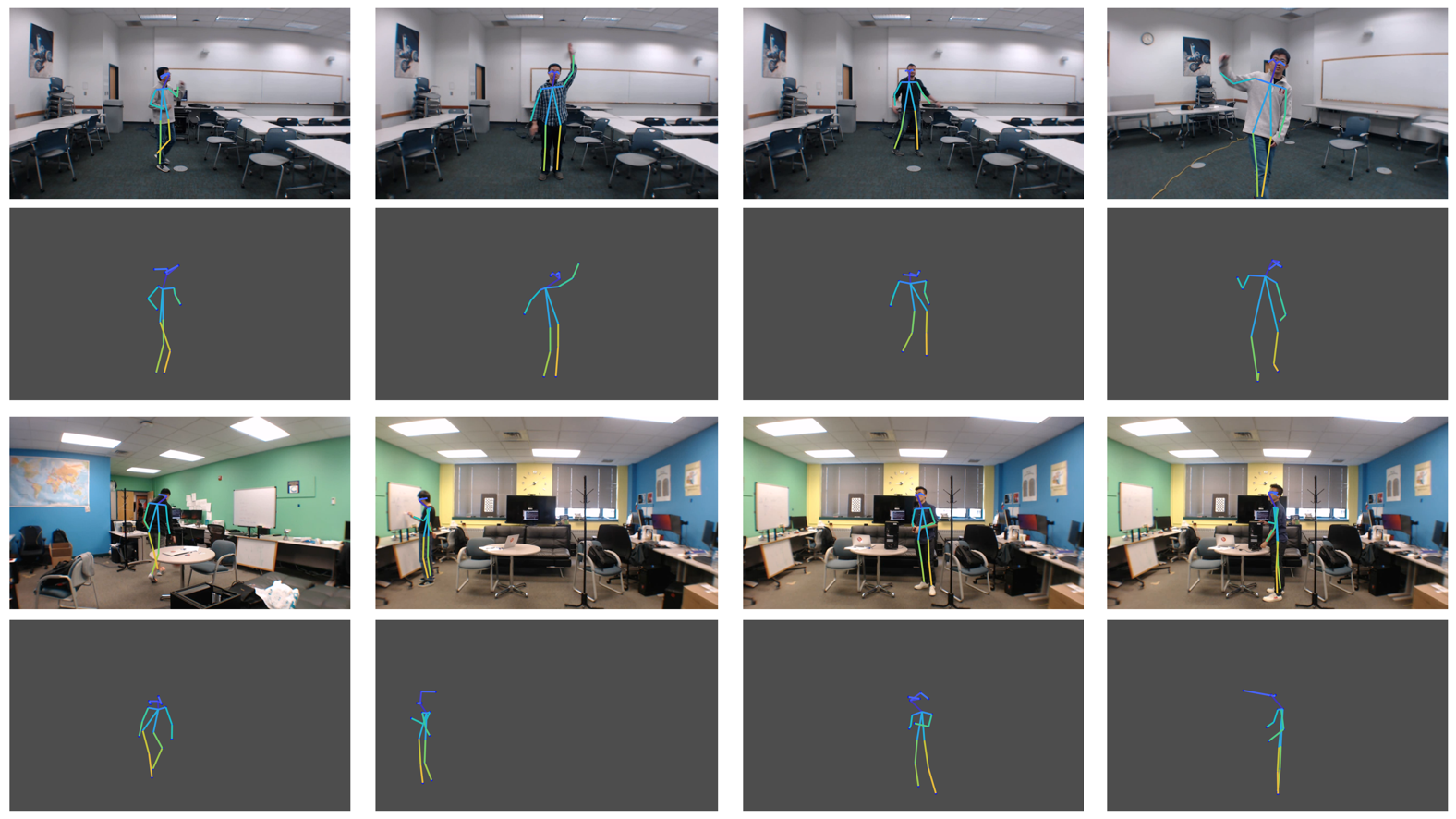}
    \caption{Result samples in 2 rooms.}
    \label{fig:result}
\end{figure}

Table.~\ref{tab:pck} shows the estimation performance of 18 body keypoint in $\mathrm{PCK}$@5,  $\mathrm{PCK}$@10, $\mathrm{PCK}$@20, $\mathrm{PCK}$@30, $\mathrm{PCK}$@40, and $\mathrm{PCK}$@50. From the table, we can see WiSPPN do pose estimation well. Figure.~\ref{fig:result} illustrates some  estimation comparisons between AlphaPose and WiSPPN. The results show that WiSPPN can work single pose estimation with comparable results to cameras.

\begin{table}[t]
\centering
\begin{tabular}{|l|l|l|l|l|l|l|l|}
\hline
Order & Keypoint    & PCK@5    & PCK@10    & PCK@20    & PCK@30  & PCK@40  & PCK@50  \\ \hline
1     & Nose        & 0.0222 & 0.1072 & 0.332  & 0.5386 & 0.6824 & 0.7634 \\ \hline
2     & Neck        & 0.0784 & 0.2222 & 0.5255 & 0.7007 & 0.8157 & 0.8797 \\ \hline
3     & R. Shoulder & 0.0575 & 0.1922 & 0.502  & 0.7098 & 0.8261 & 0.8941 \\ \hline
4     & R. Elbow    & 0.0536 & 0.1673 & 0.4235 & 0.6444 & 0.7752 & 0.8601 \\ \hline
5     & R. Wrist    & 0.0353 & 0.081  & 0.2902 & 0.5085 & 0.6745 & 0.7869 \\ \hline
6     & L. Shoulder & 0.0575 & 0.2026 & 0.4993 & 0.7111 & 0.8366 & 0.9059 \\ \hline
7     & L. Elbow    & 0.0431 & 0.1373 & 0.4131 & 0.6275 & 0.7725 & 0.8732 \\ \hline
8     & L. Wrist    & 0.0235 & 0.068  & 0.2601 & 0.4928 & 0.6405 & 0.7765 \\ \hline
9     & R. Hip      & 0.0471 & 0.1477 & 0.4497 & 0.6536 & 0.7869 & 0.8575 \\ \hline
10    & R. Knee     & 0.0418 & 0.1373 & 0.3869 & 0.583  & 0.7425 & 0.8484 \\ \hline
11    & R. Ankle    & 0.017  & 0.0771 & 0.2627 & 0.4588 & 0.5987 & 0.7085 \\ \hline
12    & L. Hip      & 0.0458 & 0.1712 & 0.4523 & 0.6484 & 0.7791 & 0.8824 \\ \hline
12    & L. Knee     & 0.0353 & 0.1216 & 0.3856 & 0.617  & 0.7843 & 0.8693 \\ \hline
14    & L. Ankle    & 0.0209 & 0.0627 & 0.2471 & 0.4497 & 0.6261 & 0.7268 \\ \hline
15    & R. Eye      & 0.0431 & 0.1477 & 0.4301 & 0.6183 & 0.7516 & 0.817  \\ \hline
16    & L. Ear      & 0.0288 & 0.1542 & 0.3778 & 0.5712 & 0.6641 & 0.7346 \\ \hline
17    & R. Ear      & 0.0405 & 0.132  & 0.3791 & 0.6039 & 0.7085 & 0.7987 \\ \hline
18    & L. Ear      & 0.0353 & 0.1281 & 0.366  & 0.5111 & 0.6288 & 0.7085 \\ \hline\hline
\multicolumn{2}{|c|}{Average} &    0.04  &  0.14  &  0.38  &  0.59 &   0.73 &   0.82  \\ \hline
\end{tabular}
\label{tab:pck}
\caption{Results in PCK. `R.' and `L.' are for right and left respectively.}
\end{table}

% \subsection{Ablation Study}

% Baseline method: fully-connected network to regress keypoint coordinates directly. 

% \begin{table*}[t]
% \centering
% \begin{tabular}{c|cccccccccccccccccc}
% method         & \#0 & \#1 & \#2 & \#3 & \#4 & \#5 & \#6 & \#7 & \#8 & \#9 & \#10 & \#11 & \#12 & \#13 & \#14 & \#15 & \#16 & \#17 \\ \hline
% baseline  & \#0 & \#1 & \#2 & \#3 & \#4 & \#5 & \#6 & \#7 & \#8 & \#9 & \#10 & \#11 & \#12 & \#13 & \#14 & \#15 & \#16 & \#17 \\
% WiSPPN+PAM  & \#0 & \#1 & \#2 & \#3 & \#4 & \#5 & \#6 & \#7 & \#8 & \#9 & \#10 & \#11 & \#12 & \#13 & \#14 & \#15 & \#16 & \#17 \\
% WiSPPN+PAM+OKS & \#0 & \#1 & \#2 & \#3 & \#4 & \#5 & \#6 & \#7 & \#8 & \#9 & \#10 & \#11 & \#12 & \#13 & \#14 & \#15 & \#16 & \#17
% \label{tab:ablation}
% \end{tabular}\caption{Ablation study of PAM and OKS loss on  PCK\@20.}
% \end{table*}

% \subsection{Discussion}